\documentclass{article}

\usepackage{microtype}
\usepackage{graphicx}
\usepackage{subfigure}
\usepackage{booktabs} 

\usepackage{hyperref}

\usepackage[accepted]{icml2025}

\usepackage{amsmath}
\usepackage{amssymb}
\usepackage{mathtools}
\usepackage{amsthm}
\usepackage{tabularx}

\usepackage[capitalize,noabbrev]{cleveref}

\theoremstyle{plain}

\theoremstyle{definition}

\theoremstyle{remark}

\usepackage{tabularx}
\usepackage{booktabs}
\usepackage{multirow}
\usepackage{times}
\usepackage{latexsym}
\usepackage{graphicx}    
\usepackage{subcaption}  
\usepackage{booktabs}  
\usepackage{array}     
\usepackage[export]{adjustbox}
\usepackage{algorithm}

\usepackage[textsize=tiny]{todonotes}

\icmltitlerunning{Submission for ICML 2025}

\begin{document}

\twocolumn[
\icmltitle{Fine-Tuning Vision-Language Models for Visual Navigation Assistance}


\begin{icmlauthorlist}
\icmlauthor{Xiao Li}{ufl}
\icmlauthor{Bharat Gandhi}{ufl}
\icmlauthor{Ming Zhan}{ufl}
\icmlauthor{Mohit Nehra}{ufl}
\icmlauthor{Zhicheng Zhang}{ufl}
\icmlauthor{Yuchen Sun}{ufl}
\icmlauthor{Meijia Song}{umn}
\icmlauthor{Naisheng Zhang}{nyu}
\icmlauthor{Xi Wang}{ufl}
\end{icmlauthorlist}

\icmlaffiliation{ufl}{University of Florida, Gainesville, FL, USA}
\icmlaffiliation{umn}{University of Minnesota, Minneapolis, MN, USA}
\icmlaffiliation{nyu}{New York University, New York, NY, USA}

\icmlcorrespondingauthor{Xiao Li}{xiao.li@ufl.edu}

\icmlkeywords{Visual Navigation, Vision Language Model, LLM Fine-tuning}

\vskip 0.3in
]




\printAffiliationsAndNotice{}

\begin{abstract}
We address vision-language-driven indoor navigation to assist visually impaired individuals in reaching a target location using images and natural language guidance. Traditional navigation systems are ineffective indoors due to the lack of precise location data. Our approach integrates vision and language models to generate step-by-step navigational instructions, enhancing accessibility and independence. We fine-tune the BLIP-2 model with Low Rank Adaptation (LoRA) on a manually annotated indoor navigation dataset. We propose an evaluation metric that refines the BERT F1 score by emphasizing ‘directional’ and ‘sequential’ variables, providing a more comprehensive measure of navigational performance. After applying LoRA, the model significantly improved in generating directional instructions, overcoming limitations in the original BLIP-2 model. 

\end{abstract}

\section{Introduction}
The challenge of indoor navigation for visually impaired individuals has long been hindered by the limitations of traditional navigation systems, which rely on GPS and predefined maps that are ineffective in indoor environments \cite{Anderson2018RoomToRoom}. Without precise location data, these systems cannot provide the necessary assistance in navigating unfamiliar spaces. To address this, we propose a vision-language-driven model designed to generate step-by-step navigational instructions based on visual inputs and natural language instructions. By integrating cutting-edge vision and language models, our approach enables users to better understand their surroundings and receive intuitive, language-based directions, significantly improving accessibility and independence.

\begin{figure}[ht]
    \centering
    \includegraphics[width=0.5\textwidth]{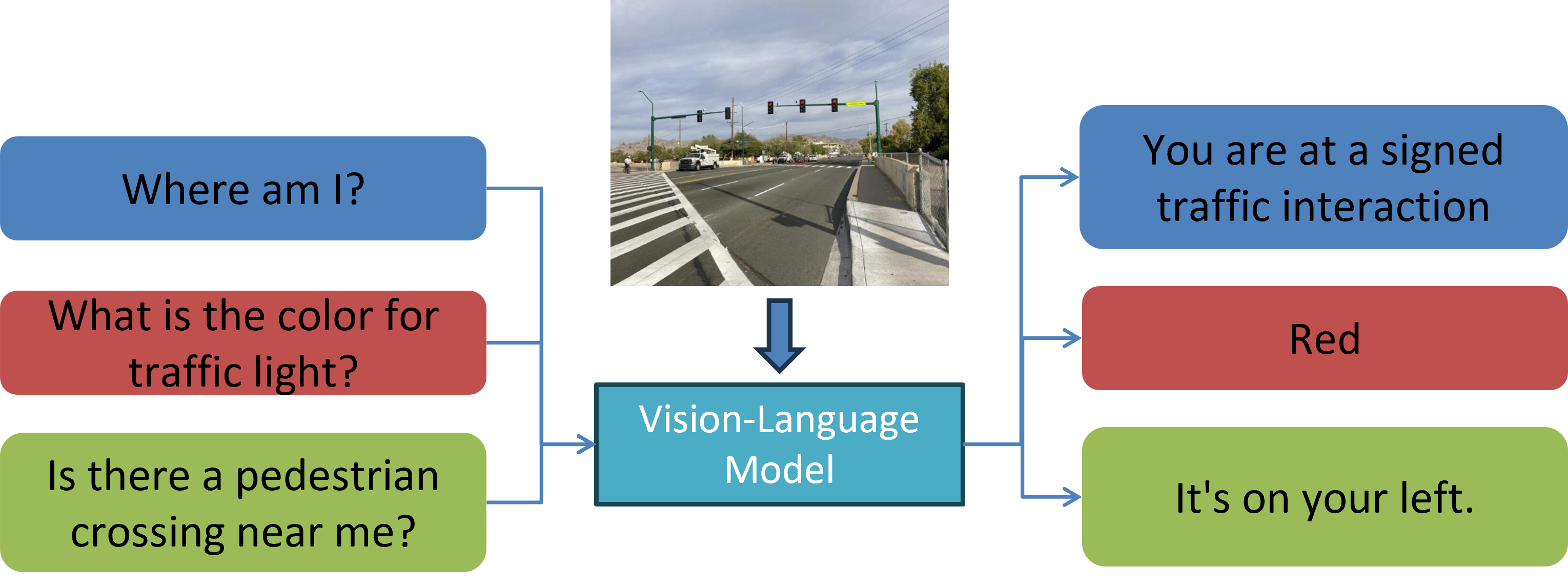}
    \caption{An example of vision language model applicatioon.}
    \label{fig:example}
\end{figure}

Our solution leverages a Vision-Language Model (VLM), specifically the BLIP-2 model \cite{Li2023BLIP2}, fine-tuned using Low Rank Adaptation (LoRA) \cite{Hu2021LoRA} for efficient training. As shown in figure \ref{fig:example}, VLMs are capable of jointly processing a textual query and a corresponding image as inputs to generate context-aware answers. Through the collection of a unique indoor navigation dataset consisting of annotated images and corresponding question-answer pairs, the model is trained to generate accurate navigation instructions. This enables visually impaired individuals to navigate indoor spaces more easily, with real-time, context-aware guidance. This work not only enhances the BLIP-2 model's capabilities for indoor navigation tasks but also contributes to the integration of vision and language models for assistive technologies. 

The contribution of this work include 3 parts:

\begin{itemize}
    \item We create a navigation dataset specifically designed for indoor space vision queries and corresponding answers.
    \item We fine-tune the BLIP2 vision-language model using LoRA to optimize it for a domain-specific task.
    \item We compare the fine-tuned BLIP2 model with the original version and demonstrate the improvements in visual navigation query answering performance.
\end{itemize}


\section{Related Work}

The field of vision–language modeling has rapidly evolved, encompassing large-scale pretraining methods, specialized architectures for multimodal integration, and efficient adaptation techniques. In this section, we first outline foundational approaches to contrastive and generative pretraining, then describe the evolution of vision–language architectures—culminating in the BLIP series—before reviewing parameter-efficient fine-tuning strategies. We then turn to applications in navigation and assistive technologies, classic captioning and VQA methods.

\paragraph{Multi-Modal Pretraining Foundations}
Contrastive image–text pretraining has been the foundation for many recent vision–language models. Methods like CLIP~\cite{Radford2021CLIP} and ALIGN~\cite{Jia2021ALIGN} use large-scale contrastive objectives to align visual and textual embeddings. Subsequent works such as Florence~\cite{Yu2024Florence} and CoCa~\cite{Yu2023CoCa} extend this paradigm by combining generative and contrastive losses to improve both zero-shot and fine-tuned performance. This area of research served as motivation for us to apply Vision Language models challenging problems such as Visual Navigation.

\paragraph{Vision–Language Models and the BLIP Series}
The BLIP family introduced a unified approach for image captioning and VQA. BLIP~\cite{Li2022BLIP} first combined an image captioner with a discriminator for improved caption faithfulness. BLIP‑2~\cite{Li2023BLIP2} further decouples visual encoding from language generation via a lightweight Q‑former module paired with a frozen LLM, enabling parameter-efficient adaptation.

\paragraph{Parameter-Efficient Fine-Tuning (PEFT) Methods}
To adapt large models with minimal overhead, adapter-based methods have been proposed. Houlsby et al.~\cite{Houlsby2019Adapters} introduced trainable bottleneck layers within transformers. Prefix‑tuning~\cite{Li2021PrefixTuning} prepends trainable vectors to each layer’s input, while LoRA~\cite{Hu2021LoRA} factorizes weight updates into low-rank matrices. These approaches balance performance gains against the number of additional parameters. By taking advantage of PEFT, researchers are able to fine-tune the model despite limited computational resources.

\paragraph{Assistive Navigation}
Smartphone-based systems such as NavCog \cite{Ahmetovic2016NavCog,Miyazaki2019NavCog3} provide turn-by-turn audio instructions to blind travellers using BLE beacons and computer vision, reporting mean localisation errors under \(1.2\,\mathrm{m}\) and System-Usability-Scale (SUS) scores above~80 in studies with up to 43 visually-impaired participants. However, none of these technologies currently incorporate large language models (LLMs) for navigation tasks. Their language capabilities remain template-based or rule-driven, offering fixed or context-specific descriptions rather than dynamically reasoning about or conversing over complex navigational instructions using natural language. These limitations motivate our BLIP-2 + LoRA solution.

\paragraph{Interactive VQA and Visual Dialogue}

\textbf{ViLBERT}~\cite{vilbert2019} extends BERT to the vision--language domain with
two transformer streams linked by co-attention.
It set strong baselines on VQA and image–text retrieval,
yet requires \emph{end-to-end} pre-training of both streams on
millions of image–caption pairs, making adaptation costly.
Moreover, its dual-stream design hinders fluent natural-language
generation, limiting interactive dialogue use-cases.
Flamingo~\cite{flamingo2022}
bridges large frozen vision and language backbones with a
\emph{Perceiver-Resampler}, supporting interleaved image–text prompts
and achieving state-of-the-art few-shot VQA.
However, the strongest variant weighs $\!\sim\!80$ B parameters and
remains proprietary; fine-tuning or on-device deployment for assistive
navigation is impractical.
LLaVA~\cite{llava2023} aligns a CLIP ViT encoder with the
Vicuna LLM by \emph{visual-instruction tuning}.  Although it exhibits
impressive chat abilities, the connector MLP and language model are
fine-tuned \emph{end-to-end}, requiring billions of gradient updates
and large curated instruction datasets.
\textit{Relation to our Research: }BLIP-2~\cite{Li2023BLIP2} overcomes the above limitations with
a lightweight \emph{Q-Former} that
\emph{bootstraps} frozen vision encoders and frozen LLMs.
Only the Q-Former (and, in our work, LoRA adapters) are trained,
cutting trainable parameters by two orders of magnitude.
BLIP-2 attains higher zero-shot VQA accuracy than Flamingo while
using $54\times$ fewer parameters, and its fully open implementation
facilitates low-resource fine-tuning for vision-navigation assistance.

\paragraph{Human-Centred Evaluation Methodologies}
For Vision–Language Navigation (VLN), automatic metrics
(\textsc{ne}, \textsc{sr}, \textsc{spl}) ignore the semantic fidelity of
generated instructions, a deficiency noted by the
CFG-based diagnostic framework of \citet{finegrainedvln2024}. To bridge the gap between scalable automation and user-aligned assessment, we developed a flexible and scalable BERTScore metric, focusing on the directional similarity between predictions and ground truths rather than focusing totally on semantics of a sentence. This method takes into account the sequence of directional instructions and the directions themselves. The details of implementations can be found in section \ref{sec:eval}.

\section{Methodology}

We now discuss our method for fine tuning BLIP2 for Visual Navigation tasks. We first discuss the architectural design of BLIP2 followed by discussion on LoRA and its advantages in model training. We also take a look at the loss functions used during pre-training and then the novel evaluation metrics that was developed to capture the essence of directions in sentences.
\subsection{Vision-Language Model Selection} 
We employ {BLIP-2}, a state-of-the-art {Vision-Language Model (VLM)}, known for its efficient {visual feature extraction and natural language generation capabilities}.
BLIP-2 (Bootstrapped Language-Image Pretraining) is a modular vision-language framework that decouples visual and language processing through the integration of three key components: a vision encoder, a Querying Transformer (Q-Former), and a large language model (LLM).
The vision encoder is a pretrained Vision Transformer (ViT) that converts an input image $I$ into a sequence of visual embeddings:
\begin{equation}
\mathbf{v} = \mathrm{ViT}(I), \quad \mathbf{v} \in \mathbb{R}^{n \times d}
\end{equation}
where $n$ is the number of visual tokens and $d$ is the feature dimension. The encoder is kept frozen to preserve its general visual representation capabilities.
The Q-Former is a lightweight transformer module that bridges the modality gap between vision and language. It introduces a set of $m$ learnable query tokens $\mathbf{q}_{\text{learned}} \in \mathbb{R}^{m \times d}$ which attend to the frozen visual embeddings via cross-attention mechanisms. The output of the Q-Former is given by:
\begin{equation}
\mathbf{q}_{\text{out}} = \mathrm{Q\mbox{-}Former}(\mathbf{q}_{\text{learned}}, \mathbf{v})
\end{equation}
The Q-Former selectively extracts task-relevant features while remaining efficient due to the limited number of queries.
The processed visual representation $\mathbf{q}_{\text{out}}$ is projected into the embedding space of a frozen large language model (e.g., Flan-T5 or Vicuna). These projected tokens are prepended to a textual prompt before being passed to the LLM:
\begin{equation}
\mathrm{LLM\ Input} = [\mathbf{z}, \text{prompt}]
\end{equation}
\begin{equation}
\text{where} \quad \mathbf{z} = W \cdot \mathbf{q}_{\text{out}} + b
\end{equation}
Here, $W$ and $b$ are learnable projection parameters. The LLM then generates responses conditioned on both the visual context and the text input.

\subsection{Fine-Tuning with LoRA}  
To adapt BLIP-2 to our specific task, we employ {Low-Rank Adaptation (LoRA)}, a fine-tuning technique that significantly {reduces computational overhead}. Instead of modifying all model weights, {LoRA injects trainable low-rank matrices} into transformer layers, allowing efficient adaptation to new data while maintaining the integrity of pretrained knowledge. This technique not only {accelerates training} but also {reduces memory requirements}, making it well-suited for fine-tuning large-scale models like BLIP-2.  

\subsection{Loss Function During Fine-Tuning}  
During fine-tuning, we optimize the model using a {causal language modeling (CLM) loss}, specifically {Cross-Entropy Loss} for text generation. The objective is to minimize the difference between the model's predicted text sequence and the ground-truth responses from the dataset. The loss function is defined as:  

\begin{equation}
\mathcal{L} = - \sum_{t=1}^{T} \log P(y_t | y_{<t}, x)
\end{equation}

where:  
\begin{itemize}
    \item \( y_t \) is the ground-truth token at timestep \( t \),  
    \item \( y_{<t} \) represents the previously generated tokens,  
    \item \( x \) is the input image representation,  
    \item \( P(y_t | y_{<t}, x) \) is the model’s predicted probability of the next token.  
\end{itemize}


\subsection{Evaluation Metrics}
\label{sec:eval}

Standard NLP metrics such as BLEU and ROUGE primarily measure lexical overlap and do not accurately reflect the quality of navigation instructions, which depend heavily on directional and sequential correctness. To address this, we propose an \textbf{Enhanced BERTScore} that explicitly incorporates navigation-specific requirements. 





A directional conflict occurs when opposing terms appear in reference (R) and prediction (P) sentences:

\begin{equation}
\text{Conflict}(R,P) = \begin{cases}
1 & \text{if } \exists (d_i, d_j) \in \mathcal{D}\\
0 & \text{otherwise}
\end{cases}
\end{equation}

\noindent where, {$\mathcal{D}$} represents the set of conflicting directional pairs, {${d_i}$} represents the directional term from the reference (R), {${d_j}$} represents the directional term from the prediction (P).

Similarly, to quantify sequence discrepancies, we introduce a flow penalty. The penalty increases with the difference in the number of steps, reflecting the risk of missing or extra actions. Additionally, if the order of actions is reversed or substantially mismatched (e.g., "turn left then walk forward" vs. "walk forward then turn left"), the Enhanced Score is set to zero, indicating a critical sequence conflict.



\begin{equation*}
\begin{aligned}
\text{EnhancedScore} &= 
\underbrace{\alpha \cdot \text{Similarity}}_{\substack{\text{Weighted semantic} \\ \text{matching score}}} + 
\underbrace{\beta \cdot \text{FlowBonus}}_{\substack{\text{Directional flow} \\ \text{similarity}}} \\
&\quad + 
\underbrace{\gamma \cdot \text{SemanticSimilarity}}_{\substack{\text{Overall text} \\ \text{similarity}}} \\ 
&\quad + 
\underbrace{\text{SpecialCaseBoost}}_{\substack{\text{Functionally equivalent} \\ \text{instruction bonus}}}
\end{aligned}
\end{equation*}

The weights for each component in the Enhanced BERTScore formula ($\alpha$, $\beta$ and $\gamma$) were determined through iterative experimentation on a diverse set of navigation instruction test cases. We qualitatively analyzed the metric's sensitivity to directional conflicts, sequence errors, synonym recognition, and partial matches, and adjusted the weights to maximize alignment with human intuition and navigational correctness.

\paragraph{Final weighted Score} 
Our weighted metric combines semantic similarity with directional accuracy:
\begin{equation}
\begin{aligned}
\text{Final Score} &=
\underbrace{(1-w)}_{\text{Semantic}} \cdot \text{BERT F1} \\
&\quad + 
\underbrace{w}_{\text{Directional}} \cdot \text{Enhanced Score}
\end{aligned}
\end{equation}
with $w$ emphasizing directional precision over general semantic similarity.


\section{Experiments}

We begin by describing the dataset used, derived from \cite{quattoni2009recognizing}, and then discuss the compute availability we had and the configuration we used. The next part discusses the 4 different experimental setups that were employed in order to assess the performance of the model. We end this section by discussing the experimental results for all the 4 setups.
\subsection{Datasets}
We employed the publicly accessible Indoor Scene Recognition dataset \cite{quattoni2009recognizing}, as the foundational dataset for our navigation assistance system. This dataset encompasses 67 distinct indoor environment categories, accumulating a total of 15,620 individual images. Each category is adequately represented with at least 100 images, although the distribution varies, providing a broad coverage of diverse indoor settings, from simple corridors and waiting rooms to more complex restaurant or library interiors.

To align the dataset explicitly with the objectives of vision-language guided indoor navigation, we undertook a rigorous preprocessing strategy. This procedure involved systematically structuring image data into organized records that clearly define image paths alongside associated textual information—specifically designed questions and queries crafted for navigation assistance scenarios. This preprocessing step ensured consistency and clarity in data usage, significantly streamlining subsequent modeling tasks.

\begin{table}[htbp]
  \centering
  \small 
  \setlength{\tabcolsep}{2pt} 
  \begin{tabular}{>{\centering\arraybackslash\small}m{1.5cm} >{\small}m{2.8cm} >{\small}m{2.8cm}}
    \toprule
    \textbf{Image} & \textbf{Query} & \textbf{Answer} \\
    \midrule
    \includegraphics[width=1.5cm]{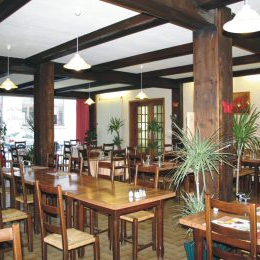} 
      & Is there any obstacle in front of me? 
      & Yes, there's a big table. \\
    \includegraphics[width=1.5cm]{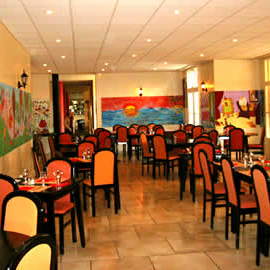} 
      & How can I exit here? 
      & Go straight for a few steps and turn left. \\
    \includegraphics[width=1.5cm]{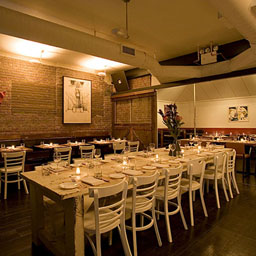} 
      & Are there any obstacles? 
      & Yes, there's a table in front of you. \\
    \includegraphics[width=1.5cm]{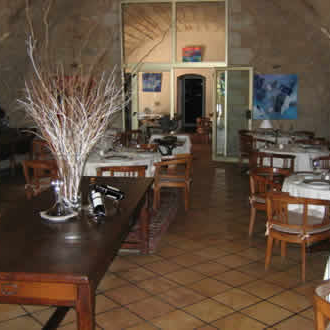} 
      & How do I leave here? 
      & Just go ahead for a few steps. \\
    \bottomrule
  \end{tabular}
  \caption{Dataset samples with manual annotations.}
  \label{tab:dataset_samples_half}
\end{table}

Complementing the preprocessing, we manually annotated nearly 1,000 image samples, generating concise and contextually relevant question-answer pairs as shown in Table \ref{tab:dataset_samples_half}. Each pair was specifically created to reflect realistic navigation scenarios encountered indoors, addressing key issues such as obstacle detection, path availability, and directional guidance. To ensure representational diversity and linguistic variability, the question-answer pairs were composed with varying phrasing while preserving semantic consistency. For example, the questions include variations like "How can I exit here?" versus "How do I leave here?", intentionally challenging models to recognize semantic equivalences across diverse expressions.

The succinctness and relevance of the question-answer pairs significantly enhance their effectiveness in real-world navigational assistance. Questions typically consist of between 4 to 8 words, while answers range from 5 to 12 words. This brevity and directness of language is crucial, ensuring clarity for users, particularly in emergency scenarios or when quick decision-making is required. Moreover, the structured simplicity of the sentences, explicitly aimed at identifying navigational pathways or obstacles, underscores the dataset’s practicality for users relying on concise auditory instructions.

To enhance dataset comprehensiveness and utility, we implemented data augmentation strategies specifically tailored to indoor navigation contexts, as demonstrated in Table \ref{tab:augmented_json_half}. For selected images, we meticulously crafted multiple distinct yet semantically consistent question-answer pairs, intentionally varying the phrasing and syntactic structures. This approach not only increased the volume of training data but also exposed the model to diverse linguistic expressions that convey identical or similar navigational intents. For instance, a single image could prompt questions like "Are there any obstacles in front of me?", "Anything blocking my path?", and "Is my way clear?", each paired with concise answers such as "Yes, there is a table," "Table ahead," and "No, table in front." This augmentation technique significantly improves the model's ability to generalize across various linguistic scenarios, ultimately enhancing its robustness, reliability, and effectiveness in generating precise, contextually relevant navigation instructions.

\subsection{Training Configuration}

We trained our model using PyTorch on an NVIDIA A100 GPU, which provided substantial computational power and memory capacity to efficiently handle our large-scale indoor navigation dataset. The training process was configured with a batch size of 32 and a learning rate of 5e-5, which was fine-tuned for optimal convergence. We employed the Adam optimizer with default parameters for stability and efficient gradient updates. The model was trained for 30 epochs, with early stopping implemented to prevent overfitting. A learning rate scheduler was used to decrease the learning rate during training to further enhance performance and generalization. The setup allowed for efficient and scalable training while maintaining high accuracy on the test set.

\begin{table}[htbp]
  \begin{minipage}[t]{0.5\textwidth}
    \raggedright
    \scriptsize
    \setlength{\tabcolsep}{4pt}
    \renewcommand{\arraystretch}{1.0}
    \begin{tabular}{>{\centering\arraybackslash}m{3cm} >{\centering\arraybackslash}m{4.6cm}}
      \toprule
      \textbf{Image} & \textbf{Augmented sample data} \\
      \midrule
      \includegraphics[valign=c,width=3.0cm]{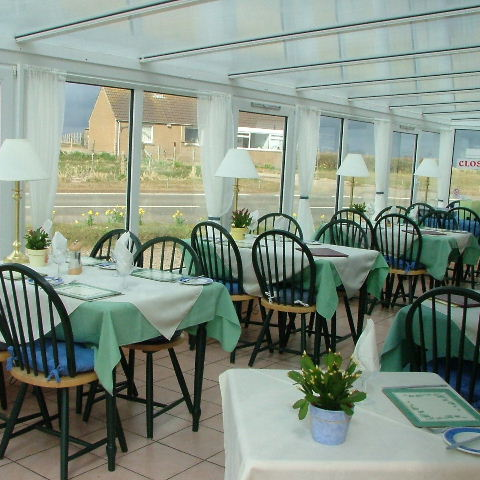} &
      \parbox[c]{\linewidth}{\scriptsize
        "0": \{\\
        \quad "path": "restaurant/bistro\_view\_of\_tables.jpg",\\
        \quad "query": "Are there any obstacles in front of me",\\
        \quad "answer": "Yes, there is a table."\\
        \},\\[0.5mm]
        "0\_1": \{\\
        \quad "path": "restaurant/bistro\_view\_of\_tables.jpg",\\
        \quad "query": "Anything blocking my path?",\\
        \quad "answer": "Table ahead."\\
        \},\\[0.5mm]
        "0\_2": \{\\
        \quad "path": "restaurant/bistro\_view\_of\_tables.jpg",\\
        \quad "query": "Is my way clear?",\\
        \quad "answer": "No, table in front."\\
        \}
      }\\[2mm]
      \midrule
      \includegraphics[valign=c,width=3.0cm]{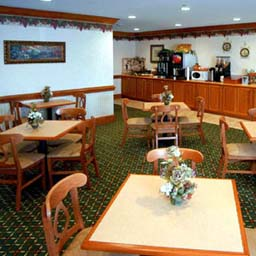} &
      \parbox[c]{\linewidth}{\scriptsize
        "1": \{\\
        \quad "path": "restaurant/food2\_450.jpg",\\
        \quad "query": "Is there anything in my way",\\
        \quad "answer": "Yes, a table is in front of you."\\
        \},\\[0.5mm]
        "1\_1": \{\\
        \quad "path": "restaurant/food2\_450.jpg",\\
        \quad "query": "Obstruction ahead?",\\
        \quad "answer": "Table in front."\\
        \},\\[0.5mm]
        "1\_2": \{\\
        \quad "path": "restaurant/food2\_450.jpg",\\
        \quad "query": "Blocked path?",\\
        \quad "answer": "Yes, table there."\\
        \}
      }\\[2mm]
      \bottomrule
    \end{tabular}
    \caption{Augmented dataset: Unique images with corresponding JSON entries.}
    \label{tab:augmented_json_half}
  \end{minipage}
\end{table}

\subsection{BLIP-2 Fine-Tuning Experiment Setups}

\begin{table*}[h]
\centering
\small
\caption{Performance Comparison. We fine-tune (FT) different parts of the Vision-Language Model (VLM) and evaluate the impact of using augmented data.}
\begin{tabularx}{\textwidth}{c|cc|c|c|c|c}
\toprule
\multirow{2}{*}{\textbf{Methods }} 
& \multicolumn{2}{c|}{\textbf{Model Fine-tuning}} 
& \multirow{2}{*}{\shortstack{\textbf{Augmented}\\\textbf{Dataset}}} 
& \multirow{2}{*}{\shortstack{\textbf{Number of}\\\textbf{Parameters Fine-tuned}}} 
& \multirow{2}{*}{\shortstack{\textbf{BERT F1}}} 
& \multirow{2}{*}{\shortstack{\textbf{Enhanced}\\\textbf{BERTScore}}} \\
\cmidrule(lr){2-3}
& \textbf{Language Model} & \textbf{Vision Model} & & & & \\
\midrule
Original BLIP2       & -- & -- & -- & -- & 0.63 & 0.46 \\
FT BLIP2 (v1)        & \checkmark & -- & -- & 9M (0.24\%)
 & 0.69 & 0.51 \\
FT BLIP2 (v2)        & \checkmark & -- & \checkmark & 9M (0.24\%) & 0.76 & 0.54 \\
FT BLIP2 (v3)        & \checkmark & \checkmark & \checkmark & 13M (0.32\%) & 0.72 & 0.52 \\
FT BLIP2 (v4)        & -- & \checkmark & \checkmark & 3.5M (0.09\%) & 0.04 & 0.11 \\
\bottomrule
\end{tabularx}
\label{tab:comparison}
\end{table*}

We conducted four distinct fine-tuning experiments to evaluate the effect of different components of the BLIP-2 architecture on downstream task performance. Each experiment involved a specific combination of finetuning the language model, the vision encoder, and the use of data augmentation. The baseline for comparison is the original BLIP-2 model without any additional training on our dataset.
The original BLIP-2 model is used without any fine-tuning. Both the vision encoder and language model remain frozen, and no additional data augmentation is used.
In the first setup, only the language model is fine-tuned, while the vision encoder remains frozen. No data augmentation is applied. Approximately 9M parameters (0.2\% of total) are updated during training.
The next experiment also fine-tunes only the language model but includes an augmented dataset to increase robustness and generalization. The vision encoder remains frozen. This setup explores the synergy between language fine-tuning and dataset diversity.
Next, both the language model and the vision encoder are fine-tuned in this configuration, along with the use of augmented data. This setup enables full adaptation of BLIP-2 to the task-specific distribution and explores the benefit of joint multi-modal tuning.
In the final setup, only the vision encoder is fine-tuned, while the language model remains frozen. The model is trained on an augmented dataset to evaluate the visual encoder's adaptability in isolation.

\subsection{Experimental Results}
\begin{enumerate}
    \item \textbf{Original BLIP-2 (No Fine-Tuning):} \\
    This configuration serves as the baseline. No components were fine-tuned, and no data augmentation was used. The model achieved a BERT F1 score of \textbf{0.63} and an Enhanced BERTScore of \textbf{0.46}.

    \item \textbf{Fine-Tuning Language Model Only:} \\
    In this setup, only the language model was fine-tuned, with 9M parameters updated (0.24\% of the full model). The vision encoder remained frozen, and no data augmentation was used. The model showed improved performance with a BERT F1 score of \textbf{0.69} and an Enhanced BERTScore of \textbf{0.51}.

    \item \textbf{Fine-Tuning Language Model with Augmented Dataset:} \\
    This experiment built on the previous configuration by introducing data augmentation. The vision encoder remained frozen, and 9M (0.24\%) parameters were fine-tuned. The performance further improved, reaching a BERT F1 score of \textbf{0.76} and an Enhanced BERTScore of \textbf{0.54}.

    \item \textbf{Joint Fine-Tuning of Language and Vision Models with Augmentation:} \\
    In this configuration, both the language and vision models were fine-tuned using an augmented dataset. The total number of trainable parameters increased to 13M (0.32\%). This led to a BERT F1 score of \textbf{0.72} and an Enhanced BERTScore of \textbf{0.52}, indicating a slight drop in F1 compared to the previous setup, though performance remained robust across both metrics.

    \item \textbf{Fine-Tuning Vision Model Only with Augmented Dataset:} \\
    In the final experiment, only the vision encoder was fine-tuned (3.5M parameters, 0.09\%), while the language model was kept frozen. Despite using data augmentation, this setup resulted in a significant performance drop, with a BERT F1 score of \textbf{0.04} and an Enhanced BERTScore of \textbf{0.11}. This suggests that tuning the vision encoder alone, without adapting the language model, severely limits model effectiveness.
\end{enumerate}

\section{Discussion}

We now discuss the experimental results and their implications, along with the limitations we encountered during the project.
\subsection{Expected Outcomes}
Prior to experimentation, we hypothesised that fine–tuning the {language model (LM)} would yield the most immediate gains because navigation instructions primarily rely on textual descriptions.  
{Joint} vision\,+\,language fine-tuning, especially with an augmented corpus, would surpass LM-only tuning by bringing visual grounding and linguistic adaptation into closer alignment.  
Fine-tuning the {vision encoder} in isolation would have limited impact, as textual decoding still governs the structure and semantics of the output.

\subsection{Key Findings}
The empirical results broadly validate but also refine these expectations:

\paragraph{LM-only tuning is highly effective.}
Updating just 0.24\% of parameters increased \textsc{BERT} F1 from~0.63 to~0.69. When coupled with data augmentation the F1 climbed to~0.76---a \textbf{21\% relative gain} over the baseline—while the Enhanced BERTScore rose to~0.54.  
\textit{Interpretation:} Most of the domain shift we face is linguistic rather than visual; the LM adapts quickly to navigation-specific phrasing once exposed to diversified trajectories and re-worded captions.

\paragraph{Data augmentation is useful.}
Across comparable configurations, augmentation contributed an additional $+0.05$ to $+0.07$ in both metrics. Synthetic re-phrasing and viewpoint jitter expanded the coverage of spatial synonyms (``ahead of'' vs.\ ``in front of'') and landmark variability, reducing over-fitting to narrow phrasings.

\paragraph{Joint tuning yields mixed returns.}
Updating both encoders (0.32\% of weights) raised the Enhanced BERTScore relative to the LM-only baseline ($0.51\!\rightarrow\!0.52$) but \emph{decreased} BERT F1 from~0.76 to~0.72.  
\textit{Interpretation:} While the vision stream benefited from exposure to augmented imagery, the additional gradient noise may have disrupted the LM’s syntactic fluency. A two-stage or lower-learning-rate schedule might stabilise this regime.

\paragraph{Vision-only tuning fails catastrophically.}
With the LM frozen, performance collapsed (F1 = 0.04). The vision encoder acquired features useful for navigation cues, but the unchanged LM could not map them into coherent descriptions. This underscores the \emph{asymmetric} dependency: language decoding can partly compensate for sub-optimal visual representations, but the reverse is not true.

\section{Limitations}

While our fine-tuning of BLIP-2 for vision-based navigation yielded promising results, there are several limitations to the current approach. One major challenge is the issue of path ambiguity. In many navigation scenarios, especially in indoor environments, multiple visually plausible routes can lead to the same destination. 

Another limitation stems from the scale and diversity of the dataset. Developing a comprehensive dataset that encompasses diverse environmental conditions is both time-consuming and resource-intensive. Our current dataset may not be sufficient for strong generalization to unseen settings. Additionally, the reliance on single-frame image inputs restricts the model's ability to incorporate temporal context, which is often essential for handling occlusions, detecting dynamic elements, and understanding directional flow in real-world navigation scenarios.

\section{Future Work}

There are several avenues for future work in our research.
First, we plan to expand our dataset to include more diverse environments, lighting conditions, and perspectives. This will enhance the model's generalization and robustness. Furthermore, we aim to extend the current framework from static image-based navigation to video-based navigation. Incorporating temporal sequences will enable the model to leverage motion cues and improve contextual understanding. In addition, future systems could benefit from integrating multimodal sensory inputs, such as inertial measurements or spatial audio cues, to help resolve visual ambiguities and ground the model's understanding in real-time physical feedback.  

\section{Conclusion}

This work presents a novel approach to indoor navigation for visually impaired individuals, utilizing a vision-language model to generate step-by-step instructions based on visual and language inputs. By integrating the BLIP-2 model with Low Rank Adaptation (LoRA), we efficiently fine-tune the model to provide accurate, real-time navigation guidance in indoor environments. Our approach enhances the model's ability to understand and respond to visual cues, significantly improving accessibility. This research advances the integration of vision and language models in assistive technologies, offering a significant contribution to improving independence and mobility for visually impaired individuals.

\bibliography{custom}
\bibliographystyle{icml2025}

\end{document}